\pdfoutput=1

\documentclass[11pt]{article}

\usepackage[final]{acl}

\usepackage{times}
\usepackage{latexsym}

\usepackage{pifont}

\usepackage[T1]{fontenc}

\usepackage[utf8]{inputenc}

\usepackage{microtype}

\usepackage{inconsolata}
\usepackage{booktabs, xcolor}
\usepackage{graphicx}
\usepackage{times}
\usepackage{soul}
\usepackage{url}
\usepackage{graphicx} 
\title{The Psychology of Falsehood: A Human-Centric Survey \\of Misinformation Detection}

\author{
 Arghodeep Nandi\textsuperscript{1},
 Megha Sundriyal\textsuperscript{2},
 Euna Mehnaz Khan\textsuperscript{3},
 Jikai Sun\textsuperscript{3},
\\
 \textbf{Emily Vraga}\textsuperscript{3},
 \textbf{Jaideep Srivastava}\textsuperscript{3},
 \textbf{Tanmoy Chakraborty}\textsuperscript{1}
\\
 \textsuperscript{1}Indian Institute of Technology Delhi,
 \textsuperscript{2}Max Planck Institute for Security and Privacy,\\
 \textsuperscript{3}University of Minnesota - Twin Cities
 \\
\textit{\{eez248395, tanchak\}@iitd.ac.in},
\textit{megha.sundriyal@mpi-sp.org},\\
\textit{\{khan0586, sun00948, ekvraga, srivasta\}@umn.edu}
}

\begin{document}
\maketitle

\begin{abstract}
Misinformation remains one of the most significant issues in the digital age. While automated fact-checking has emerged as a viable solution, most current systems are limited to evaluating factual accuracy. However, the detrimental effect of misinformation transcends simple falsehoods; it takes advantage of how individuals perceive, interpret, and emotionally react to information. This underscores the need to move beyond factuality and adopt more human-centered detection frameworks. In this survey, we explore the evolving interplay between traditional fact-checking approaches and psychological concepts such as cognitive biases, social dynamics, and emotional responses. By analyzing state-of-the-art misinformation detection systems through the lens of human psychology and behavior, we reveal critical limitations of current methods and identify opportunities for improvement. Additionally, we outline future research directions aimed at creating more robust and adaptive frameworks, such as neuro-behavioural models that integrate technological factors with the complexities of human cognition and social influence. These approaches offer promising pathways to more effectively detect and mitigate the societal harms of misinformation.


\end{abstract}

\section{Introduction}
The digital age has fundamentally transformed the way information is disseminated, leading to the rapid and widespread propagation of misinformation \citep{amoruso2020contrasting, augenstein2023factuality}. Misinformation is more than just the existence of incorrect information; it also entails complex relationships between the information and the entities that consume it. As individuals navigate this complex network of information, their perceptions and behaviours are shaped by a number of psychological and social influences \citep{ecker2022psychological}.



Misinformation is primarily classified into various categories, including fake news, misleading information, fabricated content and similar other forms \citep{altay2023survey}. However, these classifications often fail to account for the fact that even technically accurate information, when presented out of context or with bias, can also contribute to the spread of mass misperceptions. For example, the headline \textit{“A doctor dies after receiving the second dose of a vaccine”} is factually correct. However, it contributed to widespread vaccine-related misperceptions by omitting essential contextual details such as the cause of death, timing, and medical background, which allowed the narrative to exploit emotional bias and reinforce existing confirmation biases \citep{doctore_died}. Similarly, satirical news, hyper-partisan reporting, and propaganda are frequently debated as sources of misinformation \citep{patwa2021fighting, alam2022survey, altay2023survey}. To address these challenges, the focus needs to pivot from factual accuracy to comprehending how social audiences perceive and interpret information. In this study, we explore the role of perception, cognition, and social psychology in shaping the dynamics of misinformation.

\begin{table*}[!h]
\centering
\resizebox{\textwidth}{!}{%
\begin{tabular}{p{4cm}|p{4cm}|p{4cm}|p{4cm}|c}
\toprule
\textbf{Survey Paper} & \textbf{Cognitive Framing} & \textbf{Human-Centric Analysis} & \textbf{Behavioral Insights} & \textbf{Relevance to HC} \\
\midrule
\citet{guo-etal-2022-survey} & \ding{55} Not addressed & \ding{55} No focus on human factors & \ding{55} Absent & \textcolor{red}{\ding{108} Low} \\ \midrule
\citet{hardalov-etal-2022-survey}& \ding{51} Discusses stance as a proxy for belief and agreement & \ding{51} Explores how stance reflects user attitudes & \ding{115} Mentions role in misinformation spread & \textcolor{orange}{\ding{108} Moderate} \\ \midrule
\citet{akhtar-etal-2023-multimodal} & \ding{51} Notes higher credibility of multimodal misinformation & \ding{115} Highlights human susceptibility to images/videos & \ding{115} Suggests need for user studies & \textcolor{orange}{\ding{108} Moderate} \\ \midrule
\citet{vladika-matthes-2023-scientific}& \ding{55} Focuses on technical aspects & \ding{55} No discussion on user cognition & \ding{55} Absent & \textcolor{red}{\ding{108} Low} \\ \midrule
\citet{nakov-etal-2024-survey}& \ding{115} Touches on media bias perception & \ding{115} Discusses impact on public trust & \ding{115} Limited behavioral analysis & \textcolor{orange}{\ding{108} Moderate} \\ \midrule
\citet{Panchendrarajan_2024}& \ding{55} Emphasizes linguistic challenges & \ding{55} No cognitive aspects discussed & \ding{55} Absent & \textcolor{red}{\ding{108} Low} \\ \midrule
\citet{eldifrawi2024automated} & \ding{115} Emphasizes the importance of explainability in fact-checking systems & \ding{115} Discusses the need for user-understandable justifications & \ding{115} Highlights the role of user trust in automated fact-checking & \textcolor{orange}{\ding{108} Moderate} \\ \midrule
\end{tabular}%
}
\caption{Psychological and cognitive focus in misinformation survey papers. \textbf{Notation:} A cross (\ding{55}) indicates absence, a triangle (\ding{115}) signifies a tangential presence, and a tick (\ding{51}) denotes direct focus or presence. HC stands for Human Cognition.}
\label{tab:sur_paper_overview}
\end{table*}

Before the emergence of techniques addressing complex cases of misinformation, earlier studies primarily relied on relational operators to match claims with supporting evidences \citep{krishna2021proofver, Thorne18Fever, poldvere2023politifact, sundriyal2022document}. These approaches focused on evidence retrieval, effectively using evidence to train models. Advancements in Large Language Models (LLMs) have significantly reshaped the misinformation detection landscape. Recent research focuses on nuanced aspects such as ambiguity, perception, and social energy to address contemporary challenges of misinformation \citep{song2021social}. Emerging literature highlights the application of LLMs to simulate user reactions based on different user profiles and generate interaction graphs \citep{wan-etal-2024-dell}. This underscores the growing importance of graph-based algorithms in combating misinformation. The future of misinformation prevention will likely leverage Graph Neural Networks (GNNs) in conjunction with LLMs to tackle misinformation's social and psychological complexities. With the objective of better addressing the complexities of misinformation in today's context, recent studies have also started to explore these issues by developing specialised datasets with ambiguous names \citep{chiang2024merging} and persuasive contents \citep{xu-etal-2024-earth}. \citet{sundriyal2024crowd} hypothesises that the user's reactions to misinformation often reveal its accuracy. They argue that collective judgment, as expressed through public reactions, can serve as a reliable signal of the truthfulness of a given piece of information. Recently, platforms have also adopted this idea, moving toward greater reliance on community-driven tools such as community notes for content moderation \cite{borenstein2025can}. Another essential factor in disseminating misinformation is the writing style. Whether deliberate or inadvertent, the style in which misinformation is presented can significantly impact its believability. Recent studies have introduced style-agnostic training methods to reduce the impact of writing styles on misinformation detection \citep{10.1145/3637528.3671977}.

As the boundaries between AI models and psychological processes blur, there's a growing need to focus research on their intersection. This evolution signals the rise of stronger computational models that leverage data features and network structures to counter increasingly sophisticated misinformation campaigns. For instance, \citet{loth2024blessing} highlighted how Generative AI is transforming the landscape by automating the creation of misinformation (text and multimodal), effectively manipulating cognition, perception, and attitudes with minimal cost and unprecedented reach. Studies underscore the psychological impact of these advancements and the pressing need for countermeasures \citep{alam2021survey, kou2022crowd}.

As evident from Table~\ref{tab:sur_paper_overview}, various surveys on misinformation lack the required focus on cognitive framing, human-centric analysis, and behavioural insights, and their overall relevance to human cognition remains moderate to low.  In this paper, we argue for a shift toward analysing misinformation through the lens of its cognitive and psychological impacts. This transition is necessary because humans are not simply programmed machines, but conscious beings influenced by various biases and shaped by environmental and internal factors \citep{turn0search10}. Addressing these vulnerabilities requires innovative tools that integrate technology, social psychology, and insights from cognition.

\section{Classic Simplicity, But Trust it Sparingly}
\begin{figure*}[th] 
    \centering
    \includegraphics[width=\textwidth]{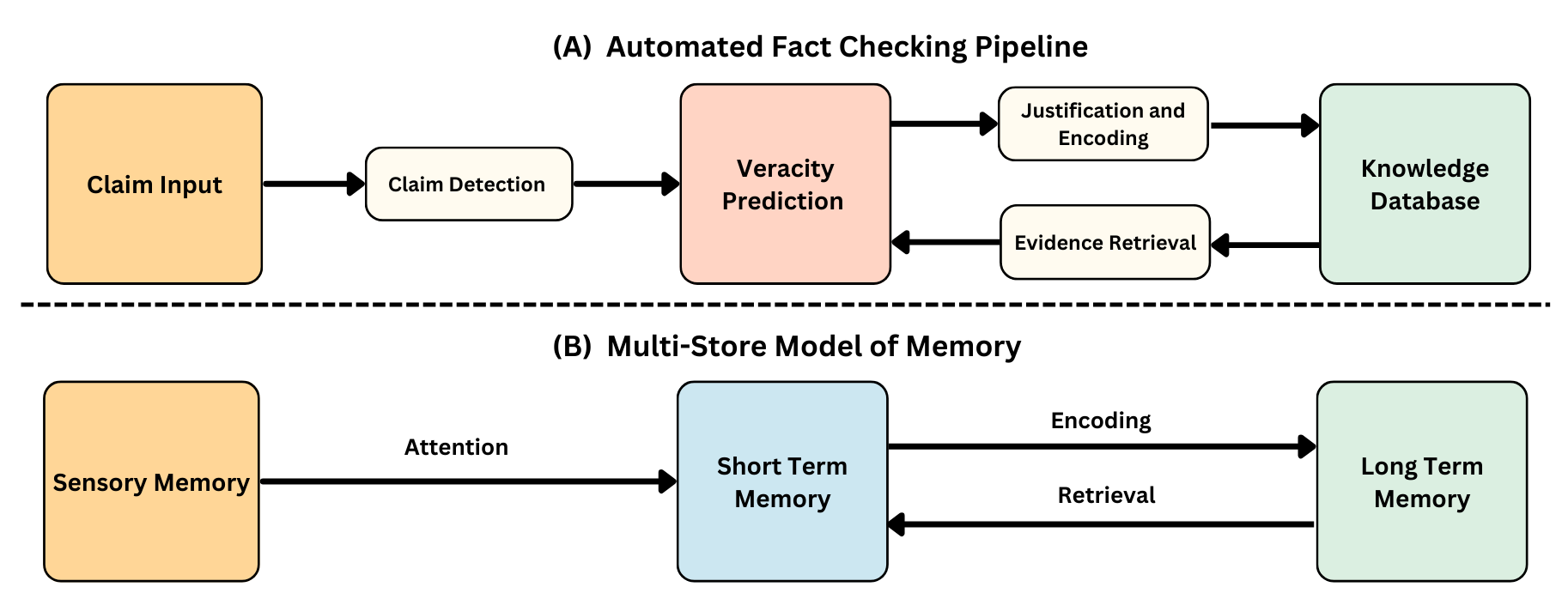} 
    \caption{A comparison between (b) Automated Fact Checking Pipeline and (b) Multi-Store Model of Memory, illustrating how misinformation detection parallels human memory processes. The fact-checking pipeline retrieves and verifies claims against a knowledge database, similar to how human cognition processes, stores, and retrieves information.}
    \label{fig:memory_model}
\end{figure*}

With the rise of the Internet, the problem of misinformation has become increasingly significant. In its early forms, misinformation often involved presenting incorrect facts in a manner that mimicked trustworthy sources. Automated Fact-Checking (AFC) modules were developed to combat this. A recent survey on AFC outlined four general stages involved in these systems \citep{eldifrawi2024automated}. Interestingly, these stages can be compared to the multi-store model of memory \citep{atkinson1968human}, which posits that memory consists of three key components: the sensory register (SR), short-term memory (STM), and long-term memory (LTM). Each AFC stage parallels memory store functions, as shown in Figure \ref{fig:memory_model}.

\paragraph{Check-worthy Claims.}
We encounter numerous stimuli daily, but not all require our attention. Only claims that are significant, verifiable, and have the potential to cause harm or mislead should be addressed \citep{wright2020claim, guo-etal-2022-survey, sundriyal2025leveraging}. The concept of claim check-worthiness can be linked to the process of attention \citep{atkinson1968human}, which facilitates the transfer of information from sensory memory to short-term memory in the multi-store memory model. A well-designed attention mechanism ensures that relevant claims are accurately identified and prioritized, enhancing the quality and reliability of the retrieved evidence or information.

\paragraph{Evidence Retrieval.}
When we receive new information, the brain compares it to what’s stored in our long-term memory. It then pulls out the memories or facts that seem to best support the idea being considered. Human intelligence is often measured by how efficiently and effectively we can retrieve and apply this stored knowledge \citep{liesefeld2016intelligence}. Similarly, the effectiveness of an AFC model is directly proportional to the performance of its evidence retrieval engine \citep{chen-etal-2024-complex, sundriyal2022document, schlichtkrull2023averitec}. Various evidence selection approaches have been discussed as classification \citep{wadden-etal-2020-fact} and regression problem; also, annotation distillation is used to mimic the annotator distribution \citep{glockner-etal-2024-ambifc}.

\paragraph{Veracity Prediction.}
Short-term memory holds the most critical information for processing, including sensory input and retrieved knowledge. Similarly, in AFC models, veracity prediction is performed using claims and the evidence that aligns with them. Both short-term memory and the AFC processing modules function as active centres where alignment, reasoning, comprehension, and classification occur in real-time \citep{BADDELEY197447}. Just as short-term memory processes and maintains information for immediate use, the AFC models rely on real-time interactions between claims and supporting evidence, dynamically updating their predictions as new data is processed. This parallel highlights the importance of adaptive, context-aware systems in both human cognition and misinformation detection models.

\paragraph{Justification Production.} 
During veracity prediction, information undergoes processing and is organized to generate justifications. This process parallels the memory encoding process, where information is summarized and structured for effective storage \citep{panigrahy2019doesmindstoreinformation}. Once encoded, the information can take the form of justifications when shared with users or become part of a knowledge base if stored in memory. Other types of models include relational models such as ProoFVer \citep{krishna2021proofver} and MultiVers \citep{wadden-etal-2022-multivers}. ProoFVer is based on the natural logic theory of compositional entailment, while MultiVers predicts rationale sentences from the evidence and classifies claim veracity using Longformer \citep{beltagy2020longformerlongdocumenttransformer} as an encoder. These models are highly explainable; however, they face significant challenges when dealing with the context and pragmatics of claims. Additionally, these models do not effectively capture complex relationships between phrases or sentences, limiting their overall performance in certain scenarios.

In the current era of multimodal content, the scope and complexity of misinformation have expanded significantly \citep{info13060284}. Misinformation now includes advanced human engineering techniques to manipulate large audiences, making evidence-retrieval-based fact-checking and relational approaches increasingly insufficient for handling its complexities. Furthermore, while foundational, the memory model under study oversimplifies and overlooks phenomena related to emotion, bias, cognition and perception \citep{article1}. Both models also fail to incorporate human behaviour in response to a claim, instead focusing solely on its content or style. Misinformation is most harmful when it has the potential to influence the masses. Additionally, filtering claims based on their potential to disrupt human behaviour is cost-effective. This strategy would enhance the efficiency of real-time fact-checking by giving importance to claims that pose large-scale risks. This highlights the need to align our research with recent advancements in psychological studies, providing a more robust framework for understanding and addressing the evolving challenges of misinformation.

In the next section, we will explore how adopting a different point of view can redefine the approach to misinformation, enabling the development of more generalizable, effective, and robust solutions.

\section{Understanding the Gestalt of Lies}
\citet{213884bd-6644-37c1-b0d4-987eea6c8f77} introduced the concept of a \textit{gestalt} to describe how an organised whole is perceived as greater than the sum of its parts. In the context of misinformation, this principle suggests that the persuasive power of falsehoods often does not lie in isolated claims but in how these claims are emotionally framed, repeated, and reinforced as a coherent narrative. Misinformation is not merely about verifying the veracity of individual claims, nor is it about assessing the truthfulness of their sum total \citep{starbird2019disinformation}. This \textit{gestalt framing} is a core mechanism behind the virality and resilience of misinformation. For instance, during the COVID-19 pandemic, individual claims (e.g., vaccines cause harm, coronavirus is a bioweapon) were either misleading or debunked when evaluated independently. Yet, when arranged into a larger, emotionally charged narrative of institutional betrayal, the gestalt effect overrides the individual debunkings. Viewers were persuaded not by any one fact, but by the cumulative structure that appears internally consistent and emotionally resonant. This further introduces the concepts of perception and \textit{`qualia'}, the subjective, individual experience of interpreting information, into the traditional approach of verifying individual claims. Qualia, in this sense, refer to how each person internally experiences a claim or narrative, colored by prior beliefs, emotions, and contextual understanding \citep{Chalmers1995-CHAFUT}. These subjective experiences can shape whether a piece of misinformation is accepted or rejected, regardless of its factual accuracy.

The gestalt nature of misinformation has been a major bottleneck in establishing a clear definition. Cognitive psychology provides models such as the multi-store model of memory \citep{atkinson1968human}, which has been compared to traditional fact-checking models. However, these cognitive models fail to capture the broader gestalt of misinformation. Current models try to capture this gestalt in fragmented ways, addressing various aspects such as claim detection \citep{gupta2021lesa, sundriyal2021desyr, sundriyal-etal-2022-empowering}, which focuses on identifying claims within a piece of content, yet often lacks the capacity to evaluate the broader context in which these claims are made. Claim simplifications aim to break down complex assertions into simpler components \citep{sundriyal2023chaos, mittal2023lost}, making it easier to assess their validity. Claim matching involves aligning detected claims with existing verified information \citep{kazemi2021claim}. Check-worthiness evaluates whether a claim is worth verifying \citep{sundriyal2023leveraging}, considering factors like relevance and potential impact. Evidence retrieval and verification are central in sourcing relevant data to support or refute claims \citep{sundriyal2022document, Thorne18Fever}. These elements, while essential to misinformation detection, often function in isolation, limiting their ability to work together in a cohesive framework.


This gap emphasises the need for misinformation interventions that go beyond factual correctness. Detection systems must account for emotional framing, contextual cohesion, and the subjective experience (or \textit{qualia}) of the information consumer. What matters is not just what is said, but how it is arranged, perceived, and interpreted collectively.


\section{Forming and Revising Beliefs}
In the battle against misinformation, understanding how humans \textit{form} and \textit{revise} their beliefs is crucial. Factual claims alone do not determine belief -- how individuals interpret and internalise these claims plays a critical role. Thus, exploring both the structure of claims and the cognitive processes involved in belief formation provides a fuller picture of misinformation’s impact.

\paragraph{Forming Beliefs.}
Two types of attitude change can be related to perceptions of the veracity of claims: incongruent change and congruent change \citep{mcguire1969}. Incongruent change occurs when a claim first believed to be `True' is altered to `False' or vice versa. Congruent change, on the other hand, involves increasing confidence in an existing label. Mathematically, these changes are typically modeled using a general loss function.

According to the psychological literature, incongruent changes are more challenging for humans than congruent changes \citep{mcguire1969}. \citet{ranadive2023specialroleclassselectiveneurons} highlighted that class-selective neurons tend to emerge within the initial few training epochs. This observation contrasts with the findings of human-based experiments, suggesting that achieving class label changes would require significantly more effort and extended training. A key limitation of this study is its focus on ResNet-50s trained on ImageNet instead of text-based data; extending the analysis to pretrained language models could offer deeper insights into class-selective neuron behavior across contexts. One possible explanation is the lack of human biases, stereotypes, and prewiring in these models, which may ease incongruent changes. Further research should include developing datasets to capture these cognitive attributes in language models.

\paragraph{Revising Beliefs.}
Both external and internal factors influence individuals’ susceptibility to misinformation and motives to disseminate it \citep{sindermann2021evaluation}. External factors relate to features of the information environment and social networks, while internal factors include personal traits and cognitive biases. Many meta-analyses have been conducted to identify the key psychological factors involved in the spread of misinformation \citep{munusamy2024psychological,nan2022people,sultan2024susceptibility}. Psychological concepts related to misinformation are detailed in Appendix \ref{appen:cog_concepts}.

As implied by the mere-exposure effect \citep{zajonc1968attitudinal}, individuals may develop positive attitudes toward information they encounter repeatedly. This contributes to a true-news bias \citep{sultan2024susceptibility}. According to confirmation bias theory, people tend to place more trust in information that aligns with their existing beliefs \citep{klayman1995varieties}. Studies further show that people who already hold a misperception are more likely to accept misinformation that confirms it, intensifying polarization in public discourse (which aligns with congruent change discussed above) \citep{zhou2022confirmation}. The heuristic-systematic model (HSM) posits that people may process information in a heuristic way, which is nonanalytic \citep{todorov2002heuristic} and makes people more easily to share misinformation \citep{sun2024heuristic}.

\section{Cognitive and Social Complexities}
Several fact-checking models have sought to break traditional barriers by incorporating human perspectives into their frameworks in recent years. \citet{chen-etal-2024-complex} highlights the challenges of evidence retrieval in real-world scenarios and emphasizes the need for a human-in-the-loop fact-checking system. This represents a larger trend in misinformation research, from rigorous, fact-based paradigms to more abstract, psychologically informed approaches. Next, we will look at significant psycho-social components that have been added into recent initiatives to counteract misinformation. A more temporal perspective can be found in Appendix \ref{appen:recent_studies}.

\paragraph{Social Energy and Perspectives.}
LLMs are limited in their ability to be used directly off-the-shelf for judging the veracity of news articles, where factual accuracy is essential. To alleviate this issue, \citet{wan-etal-2024-dell} proposed crucial steps in misinformation detection where LLMs may be introduced into the pipeline. Their model generates diverse reactions by leveraging varied user attributes and creates a user-news network using prompt-based techniques. To simulate a potential misinformation propagation process, three distinct strategies are implemented for LLMs: (i) generate a comment based on the news article, (ii) generate a comment in response to an existing comment, and (iii) select a comment for further engagement. The resulting network embodies an artificial social perspective \citep{song2021social}, which serves as the foundation for performing various tasks using GNNs. This approach integrates user interactions to enhance the model's ability to analyse and interpret complex patterns in misinformation detection. In AFaCTA \citep{ni-etal-2024-afacta}, debates between LLM agents are used as a mechanism to improve factuality assessments. During these debates, the agents engage in back-and-forth discussions, constructing reasoned arguments that explore different perspectives on a claim. This collaborative process, when combined with step-by-step fact extraction, leads to more accurate and reliable performance of the fact-checking system. 


Despite their innovation, these models face psycho-social limitations, notably the lack of interpretability in LLMs, which hinders verification of their real-world reliability. Such outputs may emerge from hallucinations \citep{guan-etal-2024-language} or biases \citep{kumar2024investigatingimplicitbiaslarge}, affecting the model’s robustness. Ensuring reliability requires metrics to assess the appropriateness and logic of these responses, improving overall stability and performance. Additionally, the scarcity of open-source LLMs raises concerns about behavioral variability, as training data differences can influence the generation of user perspectives. Prompt design also carries human biases, potentially limiting the range of outcomes. Addressing these issues requires extensive research using real-world datasets and focusing on the interpretability of models \citep{elhage2021mathematical}. Overall, the research shows us a way to simulate a misinformation propagation framework, but the data used is questionable in terms of its reliability for combating real-world misinformation.

\begin{figure*}[t]
    \centering
    \includegraphics[width=\textwidth]{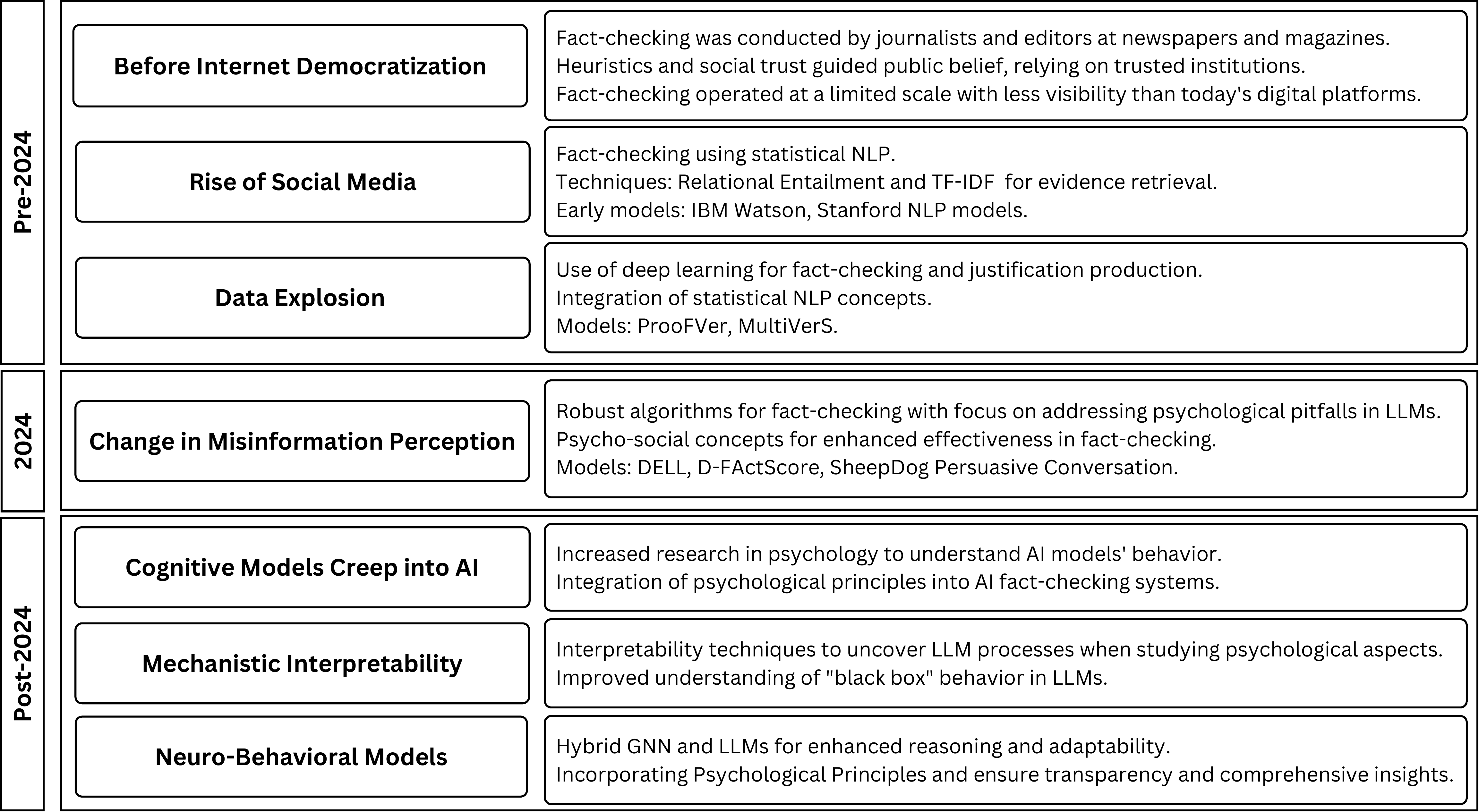}
    \caption{The timeline of the evolution of misinformation research, demonstrating significant advances in fact-checking approaches from earlier times (pre-2024), present (2024), and projected future (post-2024).}
    \label{fig:timeline}
\end{figure*}

\paragraph{Heuristics, Bias, and Challenge of Entity Ambiguity in LLMs.}
\citet{chiang2024merging} highlighted a critical issue in LLMs, where information about multiple entities is merged within the same biography, misleading users who lack prior knowledge. While the proposed D-FActScore metric seeks to address this by evaluating factuality in the presence of entity ambiguity \citep{min-etal-2023-factscore, chiang2024merging}, it treats the problem primarily as a surface-level inconsistency rather than a deeper cognitive phenomenon. As \citet{kahneman1982judgment} noted, cognitive biases influence how individuals interpret and integrate information, suggesting that entity ambiguity is not merely a technical glitch but a reflection of heuristics and mental shortcuts users employ. For example, a politically biased reader might mix up `George Soros' with unrelated people in a conspiracy article due to repeated exposure to partisan misinformation, showing how existing beliefs override facts that don't fit. In such cases, ambiguity fosters a gestalt of lies, where pieces from multiple biographies blend into a coherent but false narrative. The resulting plausibility is strengthened by the qualia of familiarity and coherence, which makes misinformation feel naturally true even when it is factually wrong. To address these deeper cognitive factors, future research should focus on improving evidence-retrieval models by developing systems that incorporate cognitive heuristics. These systems would go beyond simple search functionalities, actively mitigating biases and improving the reliability of information retrieval \citep{chaiken1989heuristic}.

Future studies could also explore the role of false cognates, words that look similar across languages but have different meanings, in spreading misinformation and hate. This area of research holds significant potential for addressing a unique and impactful dimension of the misinformation problem.

\paragraph{LLMs’ Belief Towards Misinformation via Persuasive Conversation.}
\cite{xu-etal-2024-earth} demonstrated that LLMs can shift correct beliefs when exposed to persuasive misinformation, a pattern that mirrors human susceptibility to persuasion. According to cognitive dissonance theory \citep{festinger1957theory}, when individuals hold conflicting cognitions such as knowing a fact while simultaneously encountering persuasive counterclaims, the resulting psychological discomfort motivates them to reduce the conflict by adjusting their attitudes or beliefs. The behaviour of LLMs under persuasion resembles this dissonance-reduction process since rather than maintaining an initially accurate belief, the models align their outputs with persuasive inputs in a way similar to how humans align beliefs to restore internal coherence. Psychological studies further highlight how persuasion exploits cognitive dissonance by creating internal conflict and presenting belief change as a resolution \citep{article777, book123}. Thus, the observed susceptibility of LLMs under persuasive influence parallels well-established human tendencies, reinforcing the analogy between model behaviour and cognitive dissonance.

Another significant finding is that most LLMs are susceptible to persuasive misinformation, particularly when it aligns with their prior knowledge or training data. This susceptibility mirrors confirmation bias in humans \citep{nickerson1998confirmation}. Additionally, the repetition strategy significantly increases the misinformation rate across most models. Repetition acts as a heuristic, making misinformation appear more credible without requiring systematic processing, aligning with the Heuristic-Systematic model theory \citep{chaiken1989heuristic}. In persuasive settings, consecutive misleading statements can form a gestalt of lies, where each claim builds on the previous to create a coherent narrative. This cumulative flow makes the argument seem more convincing than any single claim alone \citep{xu-etal-2024-earth}. Furthermore, LLMs exhibit sycophancy, corresponding to the theory of Attitude-Behaviour Consistency \citep{wicker1969attitudes}. While the literature uncovers valuable insights, it also highlights limitations. In humans, beliefs are shaped by more complex cognitive processes, including emotional investment, experiential memory, and subconscious biases. These deeper layers of human cognition are not addressed in this study.

Humans often resist persuasion through strategies such as counter-arguing, source skepticism, or reliance on prior knowledge \citep{brehm2015strategies}. Incorporating these resistance mechanisms into LLMs could help develop more robust strategies to counter misinformation effectively.

\paragraph{Style-based Stereotypes.}
\citet{10.1145/3637528.3671977} highlighted style manipulation using LLMs as a significant challenge to misinformation detection. It reveals that fake news camouflaged with LLM-generated styles substantially reduces state-of-the-art text-based detectors' effectiveness. Cognitive bias, a mental shortcut that aids quick situational analysis, is reflected in these models, exhibiting stereotypes toward certain styles when predicting veracity \citep{kahneman1982judgment}. Reframed news from trusted publishers leverages their credibility as a tool for deception, where credibility of the message source strongly influences compliance and belief \citep{article44}. Humans evaluate persuasive content using either the central route, which relies on logical reasoning, or the peripheral route, focusing on stylistic cues \citep{article}. Stylistic manipulation takes advantage of the peripheral route, bypassing critical analysis and enhancing the effectiveness of misinformation detection. Style cues can invoke the qualia of trust, similar to the feeling readers associate with reliable sources. This similarity makes misinformation seem credible by evoking the same subjective experience of trustworthiness.

However, several areas require further research. Emotional influence, combined with stylistic factors, could provide deeper insights. Investigating individual differences in perceiving styles as trustworthy could lead to developing robust and adaptable models. Additionally, the current evaluation framework focuses only on individual interactions with content, overlooking the social amplification of misinformation. Comparing and contrasting the effects of style and social influence could offer valuable insights into how these factors collectively and interactively shape belief in misinformation \citep{song2021social}.

\section{Future Scope and Directions}
The findings discussed in this work provide a foundational basis for future research in the field of misinformation, as illustrated in Figure \ref{fig:timeline}. However, several areas still warrant further exploration. This section outlines potential future research directions, focusing on expanding current methodologies and exploring novel approaches.

\paragraph{Datasets Unlocking Psychological Biases.}
Recent datasets such as FARM \citep{xu-etal-2024-earth} and AmbigBio \citep{chiang2024merging}, fall short in addressing the complexity of multiple psychological biases simultaneously. Given that psychological biases, emotions, and perceptions are intricately linked and context-dependent, there is a clear need for datasets that better account for these intertwined factors, particularly in the context of misinformation. Creating such datasets will require extensive text annotations, necessitating collaboration between experts in linguistics and psychology. These datasets could enable the training of advanced models, including LLMs, to recognize \citep{lin-etal-2024-detection} and address these biases effectively. However, the impact of training LLMs on such datasets on their downstream task performance remains uncertain. Investigating this aspect could provide valuable insights into the development of artificial general intelligence. This might also facilitate the downgrading of certain capabilities or offer moral and emotional reasoning in LLMs, ensuring their efficiency and making them safer to deploy.


\paragraph{Cognitive Modules for LLMs.}

Mechanistic interpretability \citep{elhage2021mathematical} and techniques like LoRA (Low Rank Adaptation) and Adapters offer a promising avenue for progress by modeling weight changes and creating specific role-based modules for LLMs. This could help integrate psychological modules into LLMs without the need for additional training or fine-tuning, thereby reducing the repetitive reliance on human labour and saving significant time and compute. At the same time, LLMs remain statistical models that are prone to misgeneralization and shortcut learning. The development of datasets that capture inherent psychological factors—particularly those involving human participants—would be crucial for enabling models to reflect nuanced cognitive phenomena more faithfully. Mechanistic interpretability techniques are still in their early stages, and the integration of cognitive modules into LLMs remains a speculative yet promising direction. These modules are not limited to fact-checking but also can be used for various other applications like reasoning, cognitive neuroscience, world models, and many more.

Recent studies have demonstrated that mechanistic interpretability can be applied effectively to investigate social biases. For example, studies \citep{yu2025understandingmitigatinggenderbias, chandna2025dissectingbiasllmsmechanistic, NEURIPS2020_92650b2e} demonstrate that gender bias can be analysed in a controlled and experimentally tractable manner, such as by systematically swapping terms like “man” and “woman” in prompts. In contrast, the identification and modelling of cognitive biases remain less tractable, owing both to the scarcity of relevant datasets and to the inherent subjectivity of human cognition. While the current body of work highlights important progress, it also underscores that practical development of cognitive modules is still far from realization. A possible future direction is to train adapters on cognitive-bias datasets, once such resources become available, thereby creating reusable cognitive modules that extend beyond fact-checking to applications in reasoning, cognitive neuroscience, and world modelling. Such advancements would pave the way for more nuanced, ethical, and effective AI systems.


\begin{figure}[!t]
    \centering    \includegraphics[width=\columnwidth]{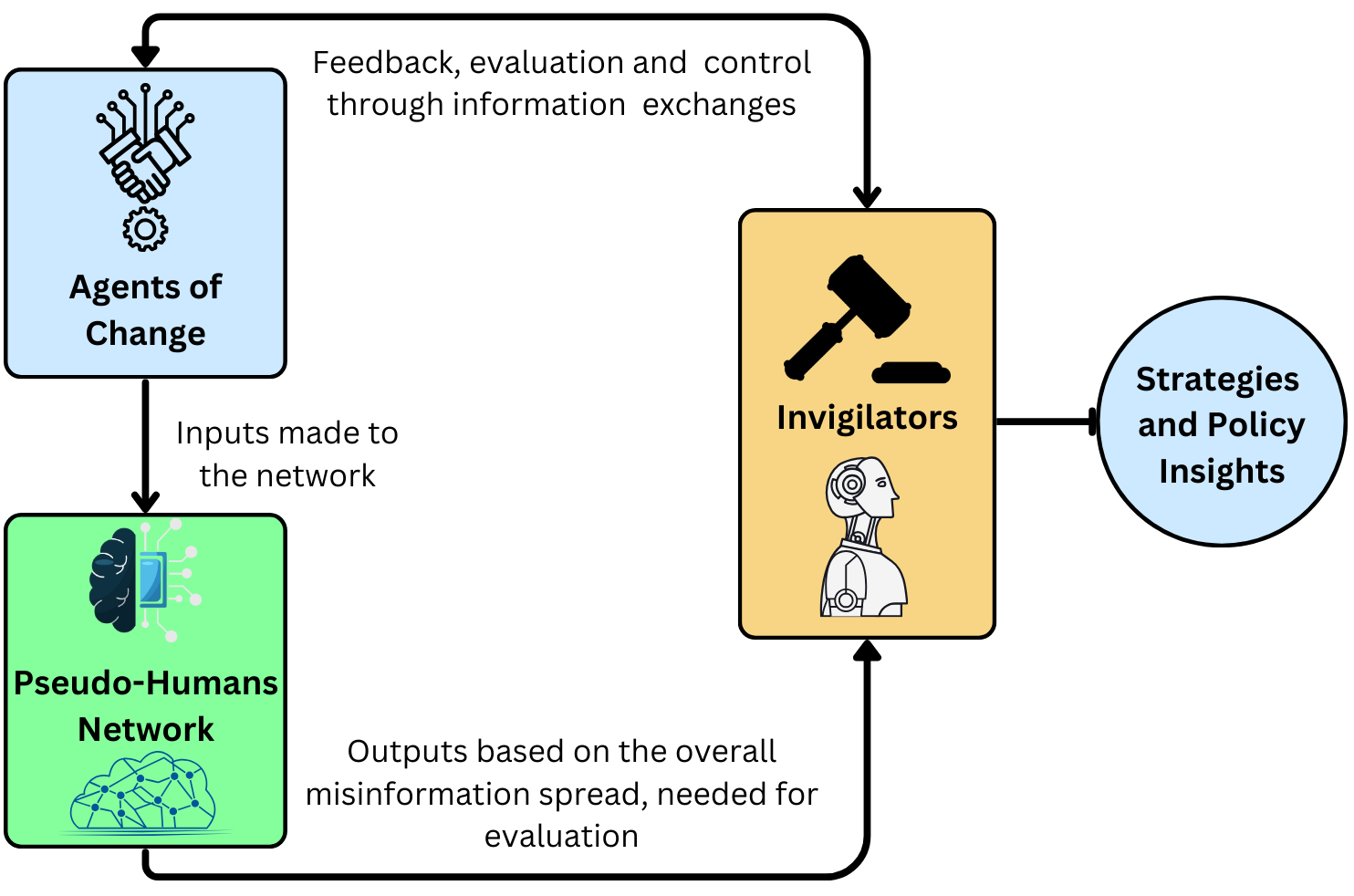}
    \caption{An illustration of the neuro-behavioural framework, showing the interaction between the three main components: agents of change, pseudo-humans network, and invigilators.}
    \label{fig:neuro_model}
\end{figure}

\paragraph{Neuro-Behavioural Models.}
World models have recently gained significant attention due to their ability to simulate the interaction between an agent and its environment \citep{https://doi.org/10.5281/zenodo.1207631}. These models create a latent representation of the world, enabling them to predict environmental dynamics by integrating perception, memory, and decision-making. By simulating scenarios internally, world models allow agents to evaluate potential actions without direct interaction, making them effective for planning and problem-solving. Multi-agent systems are essential for addressing social problems, as they interact not only with the environment but also with one another \citep{guo2024largelanguagemodelbased}. Neuro-Behavioural Models conceptualize social interactions within a simulated world. These models feature three primary types of agents, as illustrated in Figure \ref{fig:neuro_model}:
$\triangleright$ \textbf{Pseudo-humans}: Large models simulating human attributes, including perception, biases, and cognitive frameworks, with varying bias proportions for diversity. 
$\triangleright$ \textbf{Agents of Change}: Models interacting with pseudo-humans, providing inputs, analyzing outputs, and simulating scenarios.
$\triangleright$ \textbf{Invigilators}: Models that continuously evaluate the network and provide feedback to the Agents of Change, enabling dynamic input adjustments. After analysis, they assist in developing new policies and strategies.

This high-level framework requires advanced techniques, such as GNN, Reinforcement Learning, and their integration with LLMs. Its design is further supported by the rapidly growing paradigm of LLMs as autonomous agents. Recent systems such as Auto-GPT \citep{Significant_Gravitas_AutoGPT}, BabyAGI \citep{nakajima2023babyagi}, and Generative Agents \citep{10.1145/3586183.3606763} demonstrate how LLM-based multi-agent systems can collaborate to perform diverse tasks. Similarly, ReAct \citep{yao2023reactsynergizingreasoningacting} and HuggingGPT \citep{NEURIPS2023_77c33e6a} illustrate the integration of reasoning and tool use within agentic workflows, while frameworks like ChatDev \citep{qian-etal-2024-chatdev} and MetaGPT \citep{hong2024metagptmetaprogrammingmultiagent} show how specialised roles can be distributed among agents to collectively solve complex problems. These developments provide concrete demonstrations of feasibility and suggest that Neuro-Behavioural Models can emerge as a viable research direction. Advances in world models could further enable Neuro-Behavioural Models to better simulate and address these complex social challenges.

\section{Conclusion}
Through this survey, we aim to examine how misinformation research has evolved from a structural perspective, such as verifying multiple facts within a claim, to a holistic perspective, where augmented biases in datasets and psychological phenomena are integrated into misinformation detection frameworks. Misinformation has never been merely a matter of factual inaccuracies. Misinformation is instead a psychological and sociological issue that exploits human perception and reaction. This survey highlights the shift in misinformation research from factuality-centric approaches to cognitively grounded frameworks. We emphasize that combating misinformation requires more than detecting falsehoods; it also requires understanding belief. As misinformation narratives prioritise coherence, repetition, and emotional appeal over factual correctness, detection systems must progress beyond claim-level assessment. Future models should treat misinformation as a narrative, not just isolated claims. Neuro-behavioural simulations, psychological databases, and cognitive modules all present promising avenues. Integrating social cognition, and human-in-the-loop evaluations is no longer optional; it is essential for developing robust, adaptive, and trustworthy AI systems in the age of misinformation. Equally important is bridging mechanistic interpretability with behavioural insights to explain why models fail under persuasive or stylistic manipulation. Advancing in this direction could enable proactive interventions, where AI not only detects misinformation but anticipates and counters its psychological influence.

\section*{Limitations}

The limitations in this survey can be summarized in the following points:
\begin{enumerate}
    \item This survey focused exclusively on English fact-checking pipelines and associated cognitive phenomena. Exploring multilingual fact-checking and the variation of cognitive biases across languages and cultures remains a valuable direction for future research.
    \item This survey does not include a taxonomy-based classification, as it aims to bridge two distinct domains—cognition and computation—in the context of fact-checking and claim veracity. Developing a meaningful taxonomy in this area requires further research and time, as continued experimentation across models is needed to establish proper classifications.
    \item This survey primarily focuses on research published after 2021 to reflect recent advances in computational techniques for fact-checking.
    \item While this study focuses on theoretical psychological concepts. Incorporating application-based findings, such as those from behavioral interventions and relating them to misinformation, presents a promising route for future research.
\end{enumerate}

\section*{Acknowledgement}
T. Chakraborty acknowledges the support of the Anusandhan National Research
Foundation (DST/INT/USA/NSF-DST/Tanmoy/P-2/2024) and Rajiv Khemani Young Faculty
Chair Professorship in Artificial Intelligence.

\bibliography{custom}

\appendix
\section{Appendix}
\subsection{Psychological Foundations and Misinformation}
\label{appen:cog_concepts}

Understanding the psychological undercurrents that shape belief formation and information processing is essential to the study of misinformation. A list of additional psychological concepts central to the theme of this paper is provided in Table~\ref{tab:cogdef}. The aim of this table is twofold: to familiarise the reader with abstract yet foundational constructs from cognitive science, and to bridge two traditionally distinct domains -- cognitive psychology and computational social science. By grounding computational models in well-established psychological theory, we aim to enhance both the interpretability and effectiveness of misinformation detection systems.

Table~\ref{tab:cogdef} expands several such constructs that have proven instrumental in explaining why individuals find misinformation persuasive, why they struggle to abandon false beliefs, and how socio-cognitive mechanisms influence the spread of inaccurate content. One of the most influential of these is Cognitive Dissonance \citep{festinger1957theory}, which refers to the psychological discomfort caused by holding conflicting beliefs. Individuals try to resolve this discomfort by adjusting their attitudes or by filtering information selectively \citep{Taber}. This process is closely linked to Confirmation Bias \citep{nickerson1998confirmation} -- the inclination to favour, interpret, and recall information that supports pre-existing beliefs. Collectively, these cognitive tendencies help explain why misinformation often remains compelling and is accepted as truth, even when confronted with opposing facts.

The Heuristic-Systematic Model \citep{chaiken1989heuristic} further expands our understanding of how people process information. It proposes two parallel routes of evaluation: a heuristic mode based on cognitive shortcuts and peripheral cues, and a systematic mode grounded in deliberate, analytical reasoning. Misinformation often thrives in heuristic conditions, exploiting superficial cues such as authority \citep{butavicius2016breachinghumanfirewallsocial}, emotion \citep{SunXie2024,Martel2020}, or repetition \citep{Fazio2015,nyhan2010when}. Relatedly, Anchoring Bias \citep{tversky1974judgment} describes how initial exposure to a specific claim—regardless of its truth value—can anchor subsequent beliefs, skewing judgment.

The Availability Heuristic \citep{tversky1973availability} provides insight into how repeated exposure or vivid examples \citep{johnson1993source} can distort our perception of truth, as people tend to judge the likelihood or credibility of events based on how easily instances come to mind. Similarly, the Framing Effect \citep{tversky1981framing} demonstrates how the same information, when presented differently (e.g., as a gain or a loss), can significantly alter decision-making and belief acceptance. Misinformation often leverages emotional framing to manipulate these biases.

At the group level, Groupthink \citep{janis1972victims} highlights the risks of conformity and suppressed dissent in tightly knit or ideologically homogeneous communities. It explains how group cohesion can impair critical evaluation and accelerate the unchecked dissemination of misinformation \citep{turner1987rediscovering,sunstein}. Additionally, Persuasion Theory \citep{mcguire1969} and the principle of Attitude-Behavior Consistency \citep{petty1995attitude} emphasize the role of effective communication and the intensity, accessibility, and situational relevance of attitudes in predicting behavioural responses to misinformation.

Together, these psychological constructs form a theoretical backbone for understanding the psychological vulnerabilities exploited by misinformation. Their inclusion in computational frameworks not only improves model performance but also strengthens the interpretability and societal relevance of misinformation detection systems. This integration captures the core aim of this paper, to harmonize algorithmic detection methods with human psychological patterns, fostering interventions against misinformation that are both psychologically insightful and ethically responsible.

\begin{table*}[!h]
\centering
\resizebox{\textwidth}{!}{%
\begin{tabular}{p{13em}|p{34em}}
\toprule
 \textbf{Psychological Concept} & \textbf{Definition} \\
\midrule
 Cognitive Dissonance \cite{festinger1957theory} &  A psychological discomfort experienced when holding two or more conflicting cognitions, leading individuals to adjust their attitudes or behaviors to reduce inconsistency. \\ \midrule
 Confirmation Bias \cite{nickerson1998confirmation} &  The tendency to search for, interpret, and recall information in a way that confirms one's preexisting beliefs or hypotheses. \\ \midrule
 Persuasion \cite{mcguire1969} &  The process by which a person’s attitudes or behavior are influenced by communication from others, often via reciprocity, authority, or social proof. \\ \midrule 
 Heuristic-Systematic Model \cite{chaiken1989heuristic} &  A model proposing two modes of information processing: heuristic (using mental shortcuts) and systematic (in-depth and analytical), affecting how persuasive messages are judged. \\ \midrule
 Anchoring Bias \cite{tversky1974judgment} &  The tendency to rely too heavily on the first piece of information encountered (the “anchor”) when making decisions. \\ \midrule
 Availability Heuristic \cite{tversky1973availability} &  A mental shortcut where individuals estimate the probability of events based on how easily examples come to mind. \\ \midrule
 Groupthink \cite{janis1972victims}  &  A mode of thinking where desire for consensus in cohesive groups leads to suppression of dissent and poor decision-making. \\ \midrule 
 Framing Effect \cite{tversky1981framing} &  A cognitive bias where individuals' decisions are influenced by the way information is presented, such as emphasizing potential gains or losses. \\ \midrule 
 Attitude-Behavior Consistency \cite{petty1995attitude} &  The degree to which a person’s attitudes predict their behavior, influenced by attitude strength, accessibility, and context. \\ \midrule

\end{tabular}%
}
\caption{Key psychological concepts relevant to misinformation and their definitions.}
\label{tab:cogdef}
\end{table*}

\subsection{Overview of Recent Literature}
\label{appen:recent_studies}

Table~\ref{tab:paper_overview} presents a comprehensive list of recent studies on misinformation and fact-checking, examined from a psychological lens. This compilation not only identifies the psychological phenomena -- either explicitly studied or implicitly embedded -- in these works but also offers a temporal perspective, highlighting how psychological framing has gained prominence in more recent studies compared to earlier efforts. As such, the table serves as a valuable compass for researchers aiming to explore the evolving intersection of psychology and computational misinformation research.

A clear pattern emerges from this landscape, that while nearly all of the surveyed works concentrate on core tasks such as misinformation detection, fact-checking, and claim structuring, only a subset actively or inactively incorporates psychological theory to enhance their methodologies or explain user susceptibility. In particular, some of these stand out for their deep integration of foundational psychological constructs, including the Framing Effect, Confirmation Bias, and Cognitive Dissonance \citep{10.1145/3637528.3671977,si-etal-2024-checkwhy,xu-etal-2024-earth,liu-etal-2024-decoding}. These works draw upon classical frameworks such as the Heuristic-Systematic Model, bridging decision-making psychological phenomena with computational fact-checking models to enrich both understanding and performance.

Other studies venture into less traditionally studied but equally impactful psychological or sociological constructs. For example, \citep{wan-etal-2024-dell} and \citep{ni-etal-2024-afacta} introduce concepts such as Social Energy and Groupthink, highlighting the cognitive dynamics at the group level that influence belief propagation and acceptance of collective misinformation. This growing focus on social cognition marks a shift from isolated user modelling to more context-aware interactional paradigms.

The application of psychological theory extends further into the detection of multimodal misinformation. \citep{gupta-etal-2022-dialfact} and \citep{da-etal-2021-edited} incorporate the priming effect and the attribute error, respectively, to unravel how different modalities, textual, visual, or combined, shape perception and credibility judgments. Similarly, \cite{chiang2024merging} and \citep{saha-srihari-2024-integrating} examine availability heuristics and group thinking to account for how cognitive shortcuts and peer influence contribute to the spread of misinformation.

However, the survey also reveals an evident disparity: several technically sophisticated studies such as \citep{pan-etal-2023-fact}, \cite{fajcik-etal-2023-claim}, and \cite{wright-etal-2022-generating}—do not explicitly consider psychological constructs, suggesting a persistent gap between computational efficacy and cognitive realism. This observation underscores the significance of this review, as it highlights the need for more integrative and interdisciplinary approaches that not only optimize detection accuracy but also deepen our understanding of why and how users engage with misinformation.

Taken together, the growing incorporation of psychological theories into misinformation research signals a paradigm shift. In the future, there is strong potential for future studies to integrate cognitive and behavioural principles more deliberately into the development and evaluation of misinformation detection systems, resulting in tools that are not only intelligent but also psychologically informed.

\begin{table*}[ht]
\centering
\resizebox{\textwidth}{!}{%
\begin{tabular}{p{5cm}|p{7cm}|c|p{6.4cm}}
\toprule
\textbf{Previous Work} & \textbf{Research Focus} & \textbf{Psy. Study} & \textbf{Psychological Phenomenon} \\ 
\midrule
\citet{10.1145/3637528.3671977} & Misinformation Detection, Fact-Checking Models & \ding{51} & Framing Effect, Cognitive Bias \\ \midrule
\citet{xu-etal-2024-earth} & Misinformation Impact, Domain-Specific Techniques & \ding{51} & Persuasion, Heuristic-Systematic Model, Confirmation Bias, Cognitive Dissonance, Attitude-Behaviour Consistency\\ \midrule
\citet{wan-etal-2024-dell} & Fact-Checking Models, Misinformation Detection & \ding{51} & Social Energy \\ \midrule
\citet{ni-etal-2024-afacta} & Claim Structuring, Misinformation Detection & \ding{51} & Social Energy, Groupthink \\ \midrule
\citet{chiang2024merging} & Claim Structuring, Misinformation Detection & \ding{51} & Cognitive Bias, Availability Heuristics \\ \midrule
\citet{saha-srihari-2024-integrating} & Misinformation Detection, Multimodal Techniques & \ding{51} & Groupthink, Availability Heuristic \\ \midrule
\citet{si-etal-2024-checkwhy} & Fact-Checking Models, Argument Structure Reasoning & \ding{51} & Heuristic-Systematic Model, Cognitive Dissonance \\ \midrule
\citet{liu-etal-2024-decoding} & Misinformation Detection, Motivated Reasoning & \ding{51} & Heuristic-Systematic Model, Cognitive Dissonance \\ \midrule
\citet{deng-etal-2024-document} & Claim Structuring, Domain-Specific Techniques & \ding{55} & N/A \\ \midrule
\citet{luo-etal-2024-misinformation} & Claim Structuring, Domain-Specific Techniques & \ding{51} & Negativity Bias \\ \midrule
\citet{pan-etal-2023-fact} & Fact-Checking Systems, Claim Structuring & \ding{55} & N/A \\ \midrule
\citet{fajcik-etal-2023-claim} & Fact-Checking Systems, Claim Structuring & \ding{55} & N/A \\ \midrule
\citet{mendes-etal-2023-human} & Misinformation Detection, Domain-Specific Techniques & \ding{51} & Anchoring Bias \\ \midrule
\citet{yue2023metaadaptdomainadaptivefewshot} & Domain-Specific Techniques, Fact-Checking Models & \ding{55} & N/A \\ \midrule
\citet{liu-etal-2023-interpretable} & Multimodal Techniques, Claim Structuring & \ding{51} & Modality Bias \\ \midrule
\citet{gupta-etal-2022-dialfact} & Multimodal Techniques, Fact-Checking Models & \ding{51} & Priming Effect \\ \midrule
\citet{thai2022relicretrievingevidenceliterary} & Claim Structuring, Multimodal Techniques & \ding{55} & N/A \\ \midrule
\citet{wright-etal-2022-generating} & Fact-Checking Models, Domain-Specific Techniques & \ding{55} & N/A \\ \midrule
\citet{ousidhoum-etal-2022-varifocal} & Claim Structuring, Fact-Checking Models & \ding{55} & N/A \\ \midrule
\citet{chen-etal-2022-generating} & Claim Structuring, Misinformation Detection & \ding{55} & N/A \\ \midrule
\citet{glockner-etal-2022-missing} & Misinformation Impact, Multimodal Techniques & \ding{51} & Confirmation Bias \\ \midrule
\citet{sundriyal-etal-2022-empowering} & Claim Structuring, Fact-Checking Models & \ding{55} & N/A \\ \midrule
\citet{jiang-wilson-2021-structurizing} & Claim Structuring, Misinformation Detection & \ding{55} & N/A \\ \midrule
\citet{kazemi-etal-2021-claim} & Domain-Specific Techniques, Claim Structuring & \ding{51} & Cultural Bias \\ \midrule
\citet{sheng-etal-2021-article} & Claim Structuring, Fact-Checking Models & \ding{55} & N/A \\ \midrule
\citet{tymoshenko-moschitti-2021-strong} & Fact-Checking Systems, Misinformation Detection & \ding{55} & N/A \\ \midrule
\citet{da-etal-2021-edited} & Multimodal Techniques, Misinformation Impact & \ding{51} & Visual Bias, Attribution Error \\ \midrule
\citet{schlichtkrull-etal-2021-joint} & Fact-Checking Models, Claim Structuring & \ding{55} & N/A \\ \midrule
\end{tabular}%
}
\caption{Overview of misinformation-related papers categorized by their research focus, with indicators of psychological phenomena studied. Psy. Study indicates whether (\ding{51}) or not (\ding{55}) the work involves the study of any psychological phenomena.}
\label{tab:paper_overview}
\end{table*}

\end{document}